\documentclass[10pt,final,journal]{IEEEtran}

{}
{}
{}
\usepackage{balance}
\usepackage{amsmath,amssymb,amsfonts}
\usepackage{algpseudocode,algorithm,algorithmicx}
\usepackage{graphicx}
\usepackage{textcomp,color}
\usepackage{xcolor}
\usepackage{subfigure}
\usepackage{float}
\usepackage{flushend}
\usepackage{soul}
\usepackage{cuted}

\newcommand{\Wb}{\mathbf{W}}

\newcommand{\Fb}{\mathbf{F}}

\newcommand{\Pb}{\mathbf{P}}
\newcommand{\Qb}{\mathbf{Q}}
\newcommand{\Ub}{\mathbf{U}}

\newcommand{\Ib}{\mathbf{I}}

\newcommand{\xb}{\boldsymbol{x}}
\newcommand{\wb}{\mathbf{w}}

\begin{document}
%
\title{Reinforcement Learning Trained Observer Control for
Bearings-Only Tracking}

\author{Branko~Ristic and Sanjeev Arulampalam
\thanks{B.~Ristic is with the School
of Engineering, RMIT University, Melbourne, VIC3000, Australia,
e-mail: branko.ristic@rmit.edu.au}
\thanks{S. Arulampalam is with Defence Science and Technology Group, Australia.}
\thanks{Manuscript received xx xx, 2000; revised xx xx, xxxx.}}

\markboth{IEEE Trans. Aerospace and Electronic Systems,~Vol.~xx, No.~x, xx~xxxx}%
{Shell \MakeLowercase{\textit{et al.}}: Bare Demo of IEEEtran.cls for IEEE Journals}

\maketitle

\begin{abstract}
This paper develops a deep reinforcement learning 
based observer control policy for autonomous bearings-only tracking 
 of a moving target. The observer manoeuvre problem 
is formulated as a belief Markov decision process, 
where the belief state is represented by the posterior 
of a cubature Kalman filter (CKF). The reward function 
is designed to address two conflicting objectives: 
minimising the absolute target position estimation error 
(Euclidean distance) and maintaining CKF estimation 
consistency (Mahalanobis distance). The reward is 
formulated as a geometric interpolation between the two 
objectives on the Pareto front, parametrised by a 
weighting factor $\beta \in [0,1]$. The policy is implemented as a deep Q-network (DQN) trained over 50,000 episodes. Performance is 
evaluated over 5,000 Monte Carlo episodes and compared 
against two baselines: the perpendicular-to-bearing 
heuristic and the D-optimal Fisher information 
maximisation criterion. The results show that the DQN 
policy at $\beta = 0.7$ achieves the best trade-off 
between accuracy and robustness: it matches the 
information-theoretic baseline on mean tracking accuracy 
while reducing the worst-case error by nearly a factor 
of ten, owing to the implicit filter-consistency 
regularisation provided by the Mahalanobis term in the 
reward.
\end{abstract}


\IEEEpeerreviewmaketitle

\section{Introduction}

Bearings-only tracking (BOT) is an important technique for passive localization and tracking of moving targets.  The problem is sequential estimation of target location, speed and heading, from noise-corrupted measurements of target
line-of-sight bearing.  Two well known difficulties feature in the BOT problem: (i) Due to nonlinear measurements, the optimal solution in the sequential
Bayesian framework results in a non-Gaussian posterior probability density function (PDF); the exact propagation of this posterior in general cannot be expressed in the closed-form \cite{ristic_beyond_2004}; (ii) The target range is unobservable, unless the sensing platform (observer, ownship) performs a suitable maneuver  \cite{lingren1978position}; the optimal maneuver depends on the relative target–sensor motion geometry.
The former difficulty above is well understood, and as a result, a plethora of approximate sequential Bayesian estimators (Bayes filters) have been developed for BOT \cite{ristic_beyond_2004, leong2013gaussian}. The latter difficulty has also been studied extensively in the past, focusing on observability criteria \cite{nardone_aidala_81,observability_22} and on enhancement the observability by choosing the appropriate sensor platform maneuver.  Maneuver decisions are carried out in the presence of uncertainty (due to partially observable target state) typically using an information-theoretic criterion (ITC), such as the Fisher information  \cite{mohler1988nonlinear,passerieux_98, oshman_99}, mutual information \cite{652323},  Renyi divergence \cite{ristic_arulampalam_12}, or Cramer-Rao lower bound \cite{hernandez2004optimal}. 

The main drawback of the traditional decision making using ITC is computational: it involves optimisation over a multi-dimensional action space. The dimension of the action space grows exponentially with the number of steps ahead (horizon).  
Recently, deep reinforcement learning (DRL) has been put forward as an effective alternative to the traditional approach to finding the optimal action in the presence of uncertainty, primarily studied in a context of search for a chemical or a radiological source \cite{liu2019double, chen2021deep,lee2025enhanced}.  Unlike the traditional ITC methods, which require explicit computation over the multi-dimensional action state,  DRL can approximate optimal policies directly from high-dimensional inputs without needing to enumerate all possible actions. Admittedly, DRL is expensive to train, and requires large number of training episodes (simulation runs). However, once the policy is learned through training, its execution  is very fast: it requires a single forward pass through a deep neural network.  

The contributions of this paper are as follows. First, we develop a DRL based observer control for BOT. Designing the reward function for training is one of the most important aspects of reinforcement learning \cite{sutton1998reinforcement}. The reward must deal with two conflicting objectives, because the primary goal of reducing the target estimation error, if carried out aggressively (in order to increase the bearings rate) can result in a catastrophic failure due to the Bayes filter divergence. Both objectives we formulate as the minimum distance between the true target and its state estimate: the first objective distance is adopted as Euclidean distance (absolute error), while the second  is the Mahalanobis distance (relative error, as a measure of estimation consistency).  We formulate the DRL reward as a weighted combination of the two objective policies on the Pareto front \cite{xu2020prediction},  giving the maneuver control  designer the flexibility to choose an operating point based on its preference. The second contribution of the paper is the performance comparison of the DRL trained observer control for BOT, against two baseline algorithms: (1) Perpendicular-to-bearing heuristic; (2) maximisation of the determinant of the Fisher information matrix (a.k.a. D-optimality). 

The paper is organised as follows. Sec. \ref{s:2} introduces the technical background with a formal problem description. Sec. \ref{s:3} presents the details of the proposed DRL solution to observer control for BOT. The numerical results with discussion are given in Sec. \ref{s:4}, with conclusions drawn in Sec. \ref{s:5}. 

\section{Background with problem formulation}
\label{s:2}

\subsection{System models}
We adopt the state-space representation, with the (unknown) target state vector at discrete-time $k=0,1,2,\dots$ defined as:
\begin{equation}
\xb^t_k = \left[\begin{matrix}x^t_k &  y^t_k & \dot{x}^t_k &
\dot{y}^t_k\end{matrix}\right]^\top,
\end{equation}
where $[x^t_k,\;y^t_k]^\top 
$ and 
$[\dot{x}^t_k,\;\dot{y}^t_k]^\top   
$ are the target position and velocity vectors in Cartesian coordinates. 
The sensing platform state vector, denoted $\xb^o_k$,  {\em is known} and correspondingly defined as $\xb^o_k = \left[\begin{matrix}x^o_k & y^o_k & \dot{x}^o_k & \dot{y}^o_k\end{matrix}\right]^\top$. Target
motion and measurement models are expressed for the {\em relative} state vector:
\begin{equation}
\xb_k := \xb^t_k - \xb^o_k =\left[\begin{matrix}x_k & \dot{x}_k & y_k &
\dot{y}_k\end{matrix}\right]^\top \label{e:rel_state}
\end{equation}
as follows.
\subsubsection{Target motion model}
We consider nearly constant velocity (CV) model to describe the evolution of $\xb_k$:
\begin{equation}
\xb_{k+1} = \Fb_k\xb_k -\Ub_{k+1,k} + \wb_k \label{e:dyn_eq}
\end{equation}
where 
\[ \Fb = \begin{bmatrix} \mathbf{I}_2 & T\mathbf{I}_2 \\ \mathbf{0}_2 & \mathbf{I}_2 \end{bmatrix} \]
is the transition matrix, 
$\Ub_{k+1,k} = \xb_{k+1}^o-\Fb_k\xb_{k}^o $ is a known
deterministic vector that accounts for the
effects of observer accelerations, 
$\wb_k \sim {\cal N} (\mathbf{0},\Qb_k)$ is zero-mean white
Gaussian process noise with covariance matrix 
\[ \mathbf{Q} = q\begin{bmatrix} 
\frac{T^3}{3}\mathbf{I}_2 & \frac{T^2}{2}\mathbf{I}_2 \\ 
\frac{T^2}{2}\mathbf{I}_2 & T\mathbf{I}_2
\end{bmatrix},\]
 $T$ is the
sampling interval and $q$ is the process noise intensity \cite{barshalom_et_al_01}.

\subsubsection{Measurement model} Bearing measurement is related to the target state at time $k$  as follows:
 \begin{equation}
 z_k = h(\xb_k)+\nu_k
 \label{e:meas_eq}
 \end{equation}
 where $\nu_k$ is a zero-mean white Gaussian noise with variance
 $\sigma_\nu^2$, independent from process noise $\wb_k$, and
 \begin{equation}
 h(\xb_k) = \mbox{atan2}(y_k,x_k) \label{e:arctan2}
 \end{equation}
 is the four-quadrant inverse tangent function, resulting in the true target bearing  at time $k$.

\subsection{Observer Control Problem}
\label{s:ocp}

The first problem to solve in the context of autonomous bearings-only tracking is the recursive state estimation.  We adopt the Bayesian estimation framework and define the problem as follows: given (a) the posterior distribution of the state at time $k-1$, denoted $p(\xb_{k-1}|z_{1:k-1})$, where $z_{1:k-1} \equiv z_1,z_2,\dots, z_{k-1}$, and (b)  the new bearings measurement $z_k$, collected at time $k$, the objective is to compute the posterior distribution at time $k$, that is $p(\xb_{k}|z_{1:k})$. The initial posterior at $k=0$, that is   $p(\xb_{0})$, is assumed known in the Bayesian formulation. This problem is {\em not} the focus of the paper - we simply adopt the widespread solution based on the cubature Kalman filter (CKF) \cite{arasaratnam2009cubature,leong2013gaussian}, summarised for completeness in Sec. \ref{s:CKF_BOT}. The CKF approximates the posterior distribution by a Gaussian probability density function (PDF), that is:   $p(\xb_{k}|z_{1:k}) \approx {\mathcal N} (\xb; \widehat{\xb}_k,\Pb_k)$ at $k=0,1,\dots $, where the mean $\widehat{\xb}_k$ is the state estimate and $\Pb_k$ is the covariance matrix at $k$.

\begin{figure}[htb]
\centerline{\includegraphics[height=6.5cm]{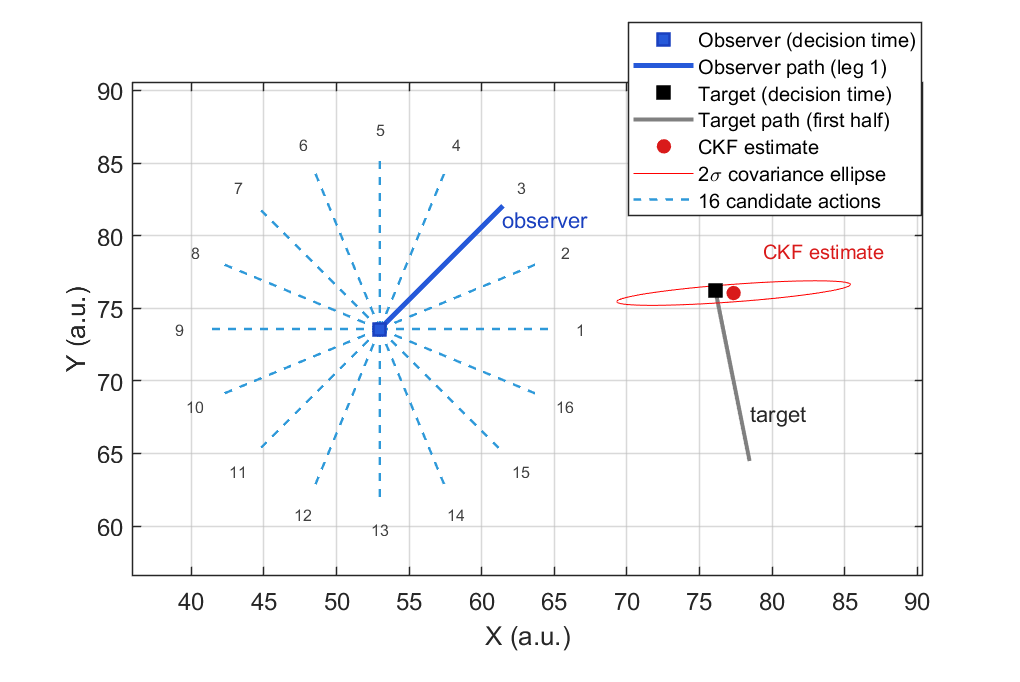}}
 \caption{\em Top-down view of an illustrative scenario. The position of the target (gray) and the observer (blue), at the time of making the decision, are marked by squares {\tiny $\Box$}. The observer has 16 options for the maneuver, indicated by the dashed lines. The target continues to move with CV. The posterior obtained by CKF at decision time is plotted in red.} \label{f:1}
\end{figure}

The focus of the paper is the second problem:  how to decide on the best observer maneuver at time $k$, based on the partial (uncertain) knowledge about the target state, expressed by the posterior  $p(\xb_{k}|z_{1:k})$. In particular, we consider the scenario from \cite{ristic_arulampalam_12}, with a two-leg observer trajectory, as shown in Fig. \ref{f:1}. Each leg consists of $M\gg 1$ time-steps.  The first leg  is plotted with a solid blue line. At the time of decision on the best maneuver, the observer position is marked by the blue square {\tiny $\Box$}. There are 16 options in Fig. \ref{f:1} for the second leg, indicated by dashed blue lines. For example, option 3 is to return to the starting position along the same path, while options 10 and 12 correspond to the change of heading by $\pm 22.5^o$. The first half of the target trajectory is shown by a solid gray line. The target position at decision time is marked with black square  {\tiny $\Box$}. The red ellipsoid in Fig. \ref{f:1} represents a contour of the CKF estimated posterior ${\mathcal N} (\xb; \widehat{\xb}_k,\Pb_k)$ at decision time. 

The heading angle of the first leg of the observer path is chosen by a simple heuristic: the path needs to be {\em perpendicular} to the first bearings measurement $z_1$. The heading angle for the second leg is the key: it must be decided autonomously by a DRL trained agent, taking into account that the relative target-observer geometry is random.

\subsection{Cubature Kalman filter for BOT}
\label{s:CKF_BOT}

\subsubsection{Equations}
This section describes one cycle of the CKF for BOT, following \cite{leong_13}.
Given the state estimate $\widehat{\xb}_{k-1}$ and the covariance matrix $\Pb_{k-1}$, the predicted state and its covariance are computed using (\ref{e:dyn_eq}) as:
\begin{eqnarray}
\widehat{\xb}_{k|k-1} & = & \Fb\, \widehat{\xb}_{k-1} - \Ub_{k,k-1} \label{e:pr1}\\
\Pb_{k|k-1} & = & \Fb\, \Pb_{k-1}\,\Fb^\top +\Qb \label{e:pr2}
\end{eqnarray}
The update step of the CKF requires to represent the predicted density $p(\xb_{k}|z_{1:k-1}) \approx \mathcal{N}(\xb_k;\widehat{\xb}_{k|k-1},\Pb_{k|k-1})$ by $L=2\times \text{dim}(\xb_k) = 8$  cubature points $\{\mathcal{X}^i_{k|k-1}\}_{1\leq i\leq L}$ with uniform weights $c_i=1/L$, $i=1,\dots,L$. The cubature points are defined as:
\begin{equation}
\mathcal{X}^i_{k|k-1} = \sqrt{\Pb_{k|k-1}}\,\xi_i + \widehat{\xb}_{k|k-1}, \hspace{2mm}i=1,\dots,L
\end{equation}
where $\xi_i$ is the $i$th column of matrix $2\left [ \mathbf{I}_{4}, -\mathbf{I}_{4}\right]$, and $\Ib_4$ is the $4\times 4$ identity matrix. The predicted measurement cubature points are computed as $\mathcal{Z}^i_{k|k-1} = h(\mathcal{X}^i_{k|k-1})$, and the predicted measurement as:
 \begin{equation}
 \widehat{z}_{k|k-1} = \frac{1}{L}\sum_{i=1}^L \mathcal{Z}^i_{k|k-1}. \label{e:pred_zhat}
  \end{equation}
  The innovation variance and the cross-covariance matrix are given by:
 \begin{eqnarray}
P^{zz}_{k|k-1} & = & \frac{1}{L} \sum_{i=1}^L (\mathcal{Z}^i_{k|k-1} - \widehat{z}_{k|k-1})^2 + \sigma_\nu^2  \label{e:Pzz} \\
\Pb^{xz}_{k|k-1} & = & \frac{1}{L} \sum_{i=1}^L (\mathcal{X}^i_{k|k-1} - \widehat{\xb}_{k|k-1})(\mathcal{Z}^i_{k|k-1} - \widehat{z}_{k|k-1}),\nonumber \\
\end{eqnarray}
respectively. Upon computing the Kalman gain
\[\Wb_k =   \frac{\Pb^{xz}_{k|k-1}} { P^{zz}_{k|k-1}},\]
 the state and covariance update are obtained using the standard Kalman filter equations \cite{ristic_beyond_2004}:
\begin{eqnarray}
 \widehat{\xb}_{k} & = &  \widehat{\xb}_{k|k-1} + \Wb_k(z_k - \widehat{z}_{k|k-1})  \label{e:cku1}\\
 \Pb_{k} & = & \Pb_{k|k-1} - \Wb_k\, P^{zz}_{k|k-1}\, \Wb_k^\top. \label{e:cku2}
 \end{eqnarray}
 
 \subsubsection{Initialisation}
 \label{sec:ckf_init}
The CKF is initialised using the first bearing measurement $z_0$ assuming a nominal range $R_0$:
\begin{equation}
    \widehat{\boldsymbol{x}}_{0} = 
    \begin{bmatrix} 
        R_0 \cos z_0 &
        R_0 \sin z_0 & 
        0 &
        0 
    \end{bmatrix}^\top.
\end{equation}
Note that the initial relative velocity is set to zero, reflecting the absence  of any prior velocity information. The initial covariance $\mathbf{P}_0$ is adopted as
\begin{equation}
    \mathbf{P}_0 = 
    \begin{bmatrix} 
        \mathbf{P}^{\boldsymbol{p}}_0 & \boldsymbol{0} \\ 
        \boldsymbol{0}                & \sigma_v^2 \mathbf{I}_2 
    \end{bmatrix}
\end{equation}
where the position covariance sub-block  $\mathbf{P}^{\boldsymbol{p}}_0$ is given by:
\begin{equation*}
    \begin{bmatrix} 
        \sigma_R^2 \cos^2 z_0 + R_0^2 \sigma_\nu^2 \sin^2 z_0 & 
        \left(\sigma_R^2 - R_0^2 \sigma_\nu^2\right) \sin z_0 \cos z_0 \\[6pt]
        \left(\sigma_R^2 - R_0^2 \sigma_\nu^2\right) \sin z_0 \cos z_0 & 
        \sigma_R^2 \sin^2 z_0 + R_0^2 \sigma_\nu^2 \cos^2 z_0
    \end{bmatrix}
\end{equation*}
which follows from  the standard first-order linearisation of the polar-to-Cartesian conversion \cite{barshalom_et_al_01}. Furthermore $\sigma_v = v_{\max}/3$ covers the maximum 
expected relative speed.

\section{Deep Reinforcement Learning Approach}
\label{s:3}

In DRL, the decision-making process is typically modeled as a Markov decision process (MDP) \texttt{-} a sequential decision making framework of interactions between the agent and its environment.  A complete, finite sequence of agent-environment interactions is referred to as an episode.  In each episode, the agent observes the environment through the  {\em observation state} $\boldsymbol{s}_k \in \mathcal{S}$ at time $k$, where $\mathcal{S}$ is the observation space (a set of all observation states), and selects an action $a_k \in \mathcal{A}$,  according to a decision-making policy. The set $\mathcal{A}$ is referred to as the action space. Then the environment transitions to the next state $\boldsymbol{s}_{k+1}$ according to probability $p(\boldsymbol{s}_{k+1}|\boldsymbol{s}_k,a_k)$ and the agent receives a reward $r_k \in \mathcal{R}$ from the environment. The reward provides a valuable feedback to the agent to improve its future decisions, in subsequent episodes. The goal of the agent is to optimise the policy by maximising the expected cumulative rewards. Through repeated episodes of agent-environment interactions, the agent learns the optimal policy.  

In MDP framework the state satisfies the Markov property — the current state contains all information necessary to predict future states and rewards. In standard DRL, the agent is assumed to have full access to this state. In the BOT context, however, the agent cannot directly observe the true target state (its position and velocity), and so the environment is effectively a partially observable MDP (POMDP). The CKF posterior serves as the belief state — a sufficient statistic summarising all past bearing measurements — and is used in place of the true state. This transforms the POMDP into a {\em belief MDP}, which satisfies the Markov property and can be treated as a standard MDP over the belief space. Next we define the elements of MDP.

\subsection{Observation state}

At each decision time   the DRL agent receives a 12-element
observation vector $\boldsymbol{s}_k \in \mathcal{S}$, constructed from the CKF estimated posterior
PDF $\mathcal{N}(\boldsymbol{x};\widehat{\boldsymbol{x}}_k, \mathbf{P}_k)$
and the most recent bearing measurement $z_k$. The observation vector is defined as:
\begin{strip}
\begin{equation}
    \boldsymbol{s}_k = \bigl[
        \widehat{\boldsymbol{p}}_k^\top, \;
        \widehat{\boldsymbol{v}}_k^\top, \;
        P_{11}, P_{12}, P_{22}, \;
        P_{33}, P_{34}, P_{44}, \;
        \cos z_k, \; \sin z_k
    \bigr]^\top \in \mathbb{R}^{12}
    \label{eq:obs}
\end{equation}
\end{strip}

\noindent where $\widehat{\boldsymbol{p}}_k = [\widehat{x}_k,\; \widehat{y}_k]^\top$ is the CKF 
relative position estimate, $\widehat{\boldsymbol{v}}_k  = [\widehat{\dot{x}}_k,\;\widehat{\dot{y}}_k]^\top$ 
is the CKF relative velocity estimate, (2 elements), $P_{ij}$ denotes
the $(i,j)$ element of $\mathbf{P}_k$, with the upper triangles
of the $2\times 2$ position and velocity sub-blocks contributing
6 elements (cross terms $P_{21} = P_{12}$ and $P_{43} = P_{34}$
are omitted by symmetry), and $(\cos z_k, \sin z_k)$ encodes
the most recent bearing as a unit vector to avoid angular
wrapping discontinuities. 

\subsection{Action}

Following the formulation in Sec. \ref{s:ocp}, the action space is discrete and consists of $N_a = 16$
candidate heading angles for the second leg of the observer
trajectory, uniformly spaced at $360^\circ/16 = 22.5^\circ$
intervals. The action space is then:
\begin{equation}
    \mathcal{A} = \left\{ a_j \;:\; \theta_j =
    (j-1)\cdot\frac{2\pi}{N_a}, \quad j = 1, \ldots, N_a \right\}.
    \label{eq:actions}
\end{equation}
The selected action $a_k \in \mathcal{A}$
determines the observer heading for all $M$ time-steps of
the second trajectory leg.

\subsection{Reward}

Designing the reward function is one of the most critical
aspects of RL \cite{sutton1998reinforcement}. In the BOT context there are
two conflicting objectives: (i)~minimising the absolute target
position estimation error (accuracy), and (ii)~maintaining
estimation consistency to prevent CKF divergence. Both are expressed in terms of the
discrepancy between the true position $\boldsymbol{p}_k$
and the CKF position estimate $\hat{\boldsymbol{p}}_k$. The
Euclidean distance measures absolute error:
\begin{equation}
    d_E = \| \hat{\boldsymbol{p}}_k - \boldsymbol{p}_k \|
    \label{eq:eucl}
\end{equation}
while the Mahalanobis distance, computed from the position
 sub-block $\mathbf{P}^{\boldsymbol{p}}_k$ of $\mathbf{P}_k$,
measures estimation consistency:
\begin{equation}
    d_M = \sqrt{
        (\hat{\boldsymbol{p}}_k - \boldsymbol{p}_k)^\top
        [\mathbf{P}^{\boldsymbol{p}}_k]^{-1}
        (\hat{\boldsymbol{p}}_k - \boldsymbol{p}_k)
    }.
    \label{eq:mah}
\end{equation}
A large Mahalanobis distance indicates that the true target
lies far outside the filter's uncertainty ellipse — a sign of
filter inconsistency or divergence. The reward is formulated
as a geometric interpolation between the two objectives
on the Pareto front \cite{xu2020prediction}:
\begin{equation}
    r_k = - d_E^{\,\beta} \cdot d_M^{1-\beta}, \qquad \beta \in [0, 1]
    \label{eq:reward}
\end{equation}
where $\beta$ is the Pareto weighting parameter. Setting
$\beta = 1$ recovers a pure Euclidean penalty (accuracy only),
while $\beta = 0$ recovers a pure Mahalanobis penalty
(consistency only); intermediate values trade off between the
two objectives. The negative sign ensures the agent maximises
reward by minimising both distances. Note that the reward is
assigned only at episode termination — intermediate rewards
are zero — reflecting the fact that tracking performance can
only be meaningfully assessed at the end of the manoeuvre.

\subsection{Episode structure}

Each episode corresponds to a single two-leg observer
trajectory. At the start of each episode, the target position
and velocity are randomly initialised, and the observer is
placed at a random position within the distance constraints
$d \in [d_{\min}, d_{\max}]$ from the target. Leg~1 is
executed deterministically inside the environment reset using
the perpendicular-to-bearing heuristic (described in
Sec.~II-B), allowing the CKF to reduce initial uncertainty
before the DRL is called upon to decide. The DRL then makes
a \emph{single} decision — the heading $a \in \mathcal{A}$ for
leg~2 — after which the environment advances for a
further $M$ time-steps, the terminal reward \eqref{eq:reward}
is computed, and the episode ends. With only one decision per
episode, no discount factor is required. The DRL agent is
trained over $N = 50,000$ episodes.

\subsection{Deep Q-Network}

The DRL agent is implemented  as a deep Q-network (DQN)
\cite{Mnih2015}, which approximates the optimal action-value
function (Q-function). 
The Q-function $Q(\boldsymbol{s}_k, a_k)$
represents the expected cumulative reward obtained by taking
action $a_k$ in observation state $\boldsymbol{s}_k$ and
following the optimal policy thereafter. 
The optimal policy
is then obtained by selecting the action that maximises the
Q-function:
\begin{equation}
    a_k^* = \arg\max_{a \in \mathcal{A}}\; Q(\boldsymbol{s}_k, a).
    \label{eq:policy}
\end{equation}
The DQN is well-suited to the problem considered in this paper because the action 
space $\mathcal{A}$ is discrete and finite, allowing the 
network to output all $N_a = 16$ Q-values simultaneously 
in a single forward pass. 

The Q-function is approximated by a fully-connected deep
neural network with parameters $\boldsymbol{\theta}$, denoted
$Q(\boldsymbol{s}_k, a; \boldsymbol{\theta})$. The network
takes the 12-element observation vector $\boldsymbol{s}_k$
as input and outputs a vector of $N_a = 16$ Q-values, one
for each candidate action. The network architecture consists
of three fully-connected hidden layers with 256, 128, and 64
neurons respectively, each followed by a ReLU activation
function.

The DQN is trained using the standard experience replay and 
reference network mechanisms \cite{Mnih2015}, with 
transitions $(\boldsymbol{s}_k, a_k, r_k, \boldsymbol{s}_{k+1})$ 
stored in a replay buffer of capacity $C = 200{,}000$ and 
sampled in mini-batches of size 256 to update the network 
parameters via the Adam optimiser with learning rate 
$\eta = 3 \times 10^{-4}$.

During training, actions are selected using an
$\varepsilon$-greedy exploration strategy: with probability
$\varepsilon$ a random action is chosen (exploration), and
with probability $1 - \varepsilon$ the action maximising the
Q-function is selected (exploitation). The exploration rate
is initialised to $\varepsilon_0 = 1.0$ and multiplied by
a decay factor of $0.99994$ after each episode, reaching
the minimum value $\varepsilon_{\min} = 0.05$ after
approximately $50{,}000$ episodes. 
The full list of DQN hyperparameters is given in
Table~\ref{tab:hyperparams}. Since each episode yields a single terminal reward with no 
intermediate rewards, the discount factor $\gamma$ has no 
effect on training; it is listed in Table~\ref{tab:hyperparams} 
for completeness.

\begin{table}[t]
\caption{DQN Hyperparameters}
\label{tab:hyperparams}
\centering
\begin{tabular}{lc}
\hline
\textbf{Hyperparameter} & \textbf{Value} \\
\hline
Hidden layer sizes      & 256, 128, 64  \\
Activation function     & ReLU          \\
Replay buffer capacity $C$       & 200,000       \\
Mini-batch size         & 256           \\
Learning rate $\eta$    & $3\times10^{-4}$ \\
Gradient clipping threshold & 1.0       \\
Reference network update frequency & 4 episodes \\
Reference network smooth factor $\tau$ & 0.01 \\
Initial exploration rate $\varepsilon_0$ & 1.0 \\
Minimum exploration rate $\varepsilon_{\min}$ & 0.05 \\
Exploration decay (per episode) & 0.99994 \\
Discount factor $\gamma$ & 0.99          \\
Training episodes $N$   & 50,000        \\
\hline
\end{tabular}
\end{table}

\section{Numerical results}
\label{s:4}

\subsection{Performance evaluation details}

The simulation parameters are summarised in 
Table~\ref{tab:simparams}. The observer and target move at 
equal speeds.
The target is initialised  
with a random constant-velocity heading, and the observer is 
placed at a random position within the distance constraints 
$d \in [d_{\min}, d_{\max}]$ from the target. Each leg of 
the observer trajectory consists of $M = 12$ time sub-steps. The 
CKF is initialised using the first bearing measurement as 
described in Sec.~\ref{sec:ckf_init}.

Performance is evaluated over $N_e = 5{,}000$ Monte Carlo 
episodes for each method. To ensure a fair comparison, 
identical random geometries are used across all methods in 
each episode, controlled via fixed random seeds. The 
following metrics are recorded at the end of each episode:
\begin{itemize}
    \item \emph{Euclidean distance} $d_E$,  eq.~(\ref{eq:eucl});
    \item \emph{Mahalanobis distance} $d_M$, eq.~\eqref{eq:mah}; 
    \item \emph{Episode reward} $r_k$, eq.~\eqref{eq:reward}: 
    the terminal reward assigned by the environment.
\end{itemize}
For each metric the mean, standard deviation, minimum and 
maximum over the $N_e$ episodes are reported.

\begin{table}[t]
\caption{Simulation Parameters}
\label{tab:simparams}
\centering
\begin{tabular}{lcc}
\hline
\textbf{Parameter} & \textbf{Symbol} & \textbf{Value} \\
\hline
Time-steps per leg               & $M$           & 12 \\
Observer/target speed       & ---           & 1 a.u./step \\
Min observer--target dist.  & $d_{\min}$    & 18 a.u. \\
Max observer--target dist.  & $d_{\max}$    & 26 a.u. \\
Bearing noise std dev       & $\sigma_\nu$  & 0.0175 rad ($\approx 1^\circ$) \\
Process noise intensity     & $q$           & $10^{-6}$ (a.u./step$^2$)$^2$ \\
Nominal initial range       & $R_0$         & 23 a.u. \\
Initial range std dev       & $\sigma_R$    & 5 a.u. \\
Max target speed (prior)    & $v_{\max}$    & 3 a.u./step \\
Evaluation episodes         & $N_e$           & 5{,}000 \\
\hline
\end{tabular}
\end{table}

\subsection{Baseline algorithms}

Two baseline observer control algorithms are used for 
comparison.

\subsubsection{Perpendicular-to-bearing (PTB) heuristic }

A necessary condition for target range observability in BOT 
is that the observer's motion has a component perpendicular 
to the line of sight \cite{nardone_aidala_81}. The simplest 
manoeuvre strategy that satisfies this condition is the 
perpendicular-to-bearing heuristic, which selects the 
heading most nearly perpendicular to the line of sight 
toward the CKF position estimate. Of the two candidate 
perpendicular directions, the one that keeps the observer 
oriented toward the predicted target position (accounting 
for estimated target velocity) is chosen. This heuristic 
requires no optimisation, but it is purely geometric and 
takes no account of the uncertainty in the target state.

\subsubsection{Information-theoretic observer (ITO)}

The second baseline is based on D-optimal Fisher 
information maximisation \cite{passerieux_98}. At the 
decision point, the algorithm evaluates all $N_a = 16$ 
candidate headings and selects the action that maximises 
the determinant of the predicted Fisher information matrix 
(FIM) accumulated over the leg:
\begin{equation}
    a^* = \arg\max_{a \in \mathcal{A}} \;
    \det \left( \mathbf{J}_k + \sum_{j=1}^{M}
    \frac{\boldsymbol{H}_j^\top \boldsymbol{H}_j}
    {\sigma_\nu^2} \right)
    \label{eq:ito}
\end{equation}
where $\mathbf{J}_k = [\mathbf{P}^{\boldsymbol{p}}_k]^{-1}$ 
is the current information matrix, and 
$\boldsymbol{H}_j = [-\Delta y_j / r_j^2,\; \Delta x_j / r_j^2]$ 
is the bearing measurement gradient evaluated at the 
predicted relative position $[\Delta x_j, \Delta y_j]^\top$ 
at time sub-step $j$, with $r_j^2 = \Delta x_j^2 + \Delta y_j^2$. 
The ITO criterion greedily maximises the information gain 
over the next leg, without accounting for filter consistency.

\subsection{Results}

\begin{figure*}[t]
    \centering
    \includegraphics[width=2\columnwidth]{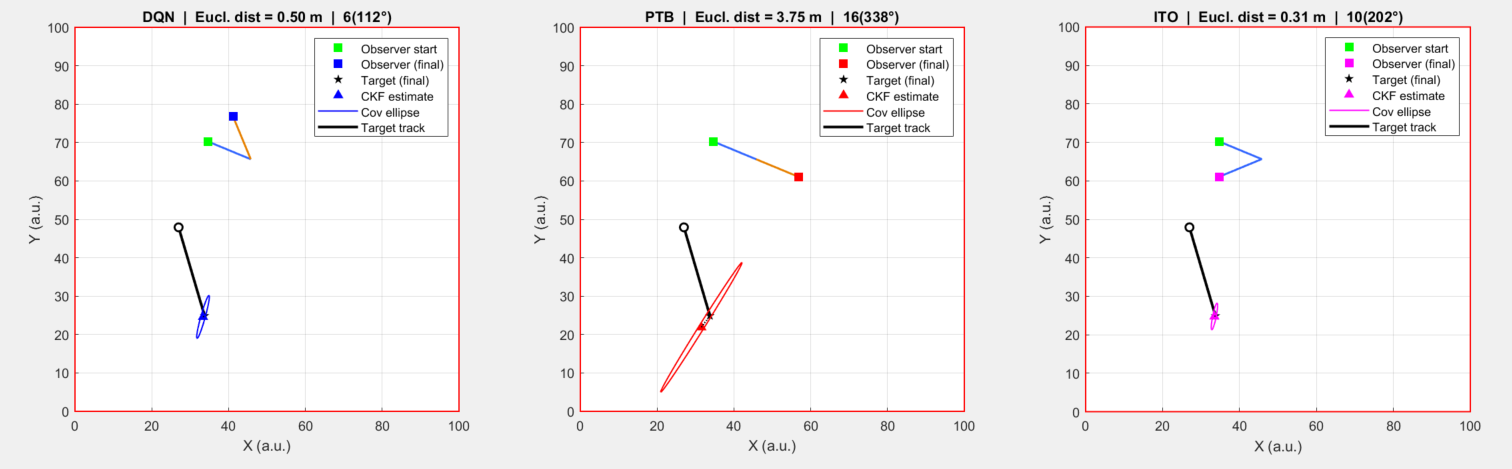}
    \caption{\em Typical single episode for the three methods 
    on the same random geometry. \emph{Left:} DQN 
    ($\beta = 0.7$), action $6$ ($112^\circ$), Euclidean 
    error $0.50$~m. \emph{Centre:} PTB heuristic, action 
    $16$ ($338^\circ$), Euclidean error $3.75$~m. 
    \emph{Right:} ITO, action $10$ ($202^\circ$), Euclidean 
    error $0.31$~m. The observer trajectory is colour-coded 
    by leg (blue: leg~1, coloured: leg~2). The open circle 
    marks the target position at the start of leg~1; the 
    black star marks the true target position at the end 
    of leg~2. The CKF posterior at episode end is shown by 
    the covariance ellipse.}
    \label{fig:single_run}
\end{figure*}

Fig.~\ref{fig:single_run} illustrates a typical single 
episode for the three methods under comparison, using the 
same random geometry. The observer starts at the green 
square and executes leg~1 using the perpendicular-to-bearing 
heuristic (identical for all three methods). At the decision 
point, the DQN ($\beta = 0.7$), PTB, and ITO policies 
select actions $6$ ($112^\circ$), $16$ ($338^\circ$), and 
$10$ ($202^\circ$), respectively, for leg~2. The target 
(black star) moves from the open circle along the solid 
black track.

The DQN policy results in the CKF 
estimate (blue triangle) at $0.50$~a.u. from the true 
target position, with a compact covariance ellipse 
indicating a well-conditioned filter. The PTB heuristic 
selects a heading that keeps the observer close to its 
starting position, resulting in a poor observation geometry 
for leg~2 and a substantially larger error of $3.75$~a.u.; 
the elongated covariance ellipse confirms that range 
uncertainty remains high. The ITO criterion achieves the 
smallest Euclidean error in this particular episode 
($0.31$~a.u.).

The summarised results obtained from $ N_e$ episodes are given in Table~\ref{tab:results}. 

\subsubsection{Effect of the Pareto weight $\beta$}

As $\beta$ increases, the reward function places 
progressively greater emphasis on minimising the Euclidean 
error relative to the Mahalanobis term. In terms of tracking 
accuracy, the mean Euclidean distance decreases from 
$2.969$~a.u. at $\beta = 0.1$ to a minimum of $2.336$~a.u. at 
$\beta = 0.7$, before rising slightly to $2.391$~a.u. at 
$\beta = 0.9$. The policy at $\beta = 0.7$ also achieves 
the lowest standard deviation among the DQN configurations 
($2.945$~a.u.), indicating the most consistent tracking 
behaviour across episodes.

At $\beta = 0.9$, however, the maximum observed error 
reaches $305.6$~m --- a twelve-fold increase relative to 
$\beta = 0.7$ --- indicating occasional catastrophic filter 
divergence when the policy aggressively pursues bearing rate 
at the expense of estimation consistency. This is precisely 
the failure mode identified in Introduction: overly aggressive 
observer manoeuvres can destabilise the CKF, leading to 
unbounded estimation error. The Mahalanobis distance remains 
broadly stable across $\beta \in \{0.1, \ldots, 0.7\}$, 
ranging between $1.256$ and $1.283$, confirming that the 
CKF maintains estimation consistency across these operating 
points. At $\beta = 0.9$, consistency degrades slightly to 
$1.317$, consistent with the increased incidence of 
divergent episodes.

\subsubsection{Comparison with baselines}

The PTB heuristic yields a mean Euclidean error of 
$3.343$~a.u. with a standard deviation of $4.445$~a.u., performing 
worse than all DQN configurations on mean accuracy. The ITO 
criterion achieves the lowest mean Euclidean error of 
$2.243$~a.u.; however, it exhibits the largest standard 
deviation ($5.951$~a.u.) and a maximum error of $246.1$~a.u., 
reflecting a tendency toward occasional severe tracking 
failures. Notably, ITO also produces the highest mean 
Mahalanobis distance of $1.557$ --- substantially above all 
DQN configurations --- indicating systematic degradation of 
filter consistency.

The DQN policy at $\beta = 0.7$ achieves a mean error of 
$2.336$~a.u., only marginally higher than ITO (by $0.093$~a.u.), 
while reducing the standard deviation by more than a factor 
of two ($2.945$~a.u. vs $5.951$~a.u.) and the maximum error by 
nearly a factor of ten ($26.2$~a.u. vs $246.1$~a.u.). This 
highlights a key advantage of the proposed reward 
formulation: by incorporating both Euclidean and Mahalanobis 
terms in the Pareto objective, the DQN learns manoeuvres 
that balance tracking accuracy with filter stability, 
avoiding the catastrophic failures exhibited by ITO.

\begin{table*}[t]
\caption{Performance comparison over $N_e = 5{,}000$ Monte 
Carlo episodes. The best value in each row is shown 
in bold.}
\label{tab:results}
\centering
\begin{tabular}{lrrrrrrr}
\hline
\textbf{Metric} 
    & $\beta=0.1$ & $\beta=0.3$ & $\beta=0.5$ 
    & $\beta=0.7$ & $\beta=0.9$ & PTB & ITO \\
\hline
Avg Euclidean dist (a.u.)  
    & 2.969 & 2.697 & 2.475 & \textbf{2.336} 
    & 2.391 & 3.343 & 2.243 \\
Std Euclidean dist (a.u.)  
    & 3.696 & 3.375 & 3.351 & \textbf{2.945} 
    & 5.288 & 4.445 & 5.951 \\
Max Euclidean dist (a.u.)  
    & 40.711 & 26.436 & 67.817 & \textbf{26.150} 
    & 305.628 & 49.037 & 246.086 \\
Avg Mahalanobis dist    
    & \textbf{1.256} & 1.258 & 1.283 & 1.267 
    & 1.317 & 1.270 & 1.557 \\
Avg episode reward$^\dagger$      
    & $-$1.299 & $-$1.402 & $-$1.562 & $-$1.750 
    & $-$2.159 & $-$1.816 & $-$1.613 \\
\hline
\multicolumn{8}{l}{$^\dagger$ DQN rewards use each agent's 
own $\beta$; PTB and ITO use $\beta = 0.5$ for reference.}\\
\end{tabular}
\end{table*}

\subsection{Discussion}

The results demonstrate that the choice of $\beta$ governs 
a clear trade-off on the Pareto front between tracking 
accuracy and filter consistency. Small values of $\beta$ 
produce conservative policies that prioritise filter health 
at the cost of accuracy, while large values produce 
aggressive policies that improve mean accuracy but risk 
catastrophic divergence. The value $\beta = 0.7$ emerges 
as the most favourable operating point, simultaneously 
achieving the lowest mean Euclidean distance, the lowest 
standard deviation, and the lowest maximum error among all 
DQN configurations.

The comparison with the two baselines reveals an important 
distinction between mean performance and robustness. The 
ITO criterion achieves a marginally lower mean Euclidean 
distance than the best DQN policy ($2.243$~m vs $2.336$~m), 
but at the cost of severe instability: its standard 
deviation is more than twice as large and its worst-case 
error exceeds $246$~m. This instability arises because ITO 
greedily maximises information gain without any mechanism 
to penalise filter divergence. In contrast, the Mahalanobis 
term in the DQN reward function acts as an implicit 
regulariser, discouraging manoeuvres that destabilise the 
CKF. The PTB heuristic, while computationally trivial, 
is consistently outperformed by all DQN configurations on 
mean accuracy, confirming that even a modest amount of 
learned policy optimisation yields meaningful gains over 
purely geometric manoeuvre selection.

From a practical standpoint, the DQN policy at $\beta = 0.7$ 
offers the best combination of accuracy and reliability for 
autonomous BOT. Once trained, the policy requires only a 
single forward pass through the neural network to produce 
a decision, making it suitable for real-time deployment. 
This contrasts with ITO, which requires evaluating the FIM 
over all candidate actions at each decision point — a 
computation that scales poorly with the number of actions 
and the planning horizon.

\section{Conclusions}
\label{s:5}

This paper has presented a DRL-based approach to observer 
control for bearings-only tracking of a moving target. 
The observer manoeuvre problem was formulated as 
a belief MDP, where the CKF posterior --- comprising the 
state estimate $\hat{\boldsymbol{x}}_k$ and covariance 
$\mathbf{P}_k$ --- serves as the belief state, 
transforming the underlying POMDP into a standard MDP 
amenable to DQN training. A novel reward function was proposed that 
balances two conflicting objectives via a geometric 
Pareto interpolation: the Euclidean distance, which 
penalises absolute tracking error, and the Mahalanobis 
distance, which penalises CKF inconsistency. The 
weighting parameter $\beta$ allows the designer to 
select an operating point on the Pareto front between 
accuracy and filter stability. The trained DQN policy was compared against two 
established baselines --- the perpendicular-to-bearing 
heuristic and the D-optimal ITO criterion --- over 
5,000 Monte Carlo episodes. The results demonstrate 
that DQN at $\beta = 0.7$ yields the most favourable results, as the best combination of accuracy and reliability for 
autonomous BOT. In addition, once trained, the DQN policy requires only a single 
forward pass through the neural network to produce a 
manoeuvre decision, making it computationally much more 
attractive than ITO for real-time deployment. 

Future work could consider extending the framework to 
multi-leg trajectories with more than one DQN decision 
per episode, incorporating target manoeuvres, or 
replacing the CKF with a particle filter to handle 
non-Gaussian posteriors more accurately.

\bibliographystyle{IEEEtran}
\bibliography{track}


\end{document}